\documentclass{article}

\usepackage{microtype}
\usepackage{graphicx}
\usepackage{subfigure}
\usepackage{booktabs} % for professional tables
\usepackage{todonotes}
\usepackage{url}

\usepackage{hyperref}

\usepackage[accepted]{icml2019}
\usepackage{flushend}

\icmltitlerunning{CADNN}

\begin{document}

\twocolumn[
\icmltitle{26ms Inference Time for ResNet-50: Towards Real-Time Execution of all DNNs on Smartphone}

% % It is OKAY to include author information, even for blind
% % submissions: the style file will automatically remove it for you
% % unless you've provided the [accepted] option to the icml2019
% % package.

% % List of affiliations: The first argument should be a (short)
% % identifier you will use later to specify author affiliations
% % Academic affiliations should list Department, University, City, Region, Country
% % Industry affiliations should list Company, City, Region, Country

% % You can specify symbols, otherwise they are numbered in order.
% % Ideally, you should not use this facility. Affiliations will be numbered
% % in order of appearance and this is the preferred way.

\begin{icmlauthorlist}
\icmlauthor{Wei Niu}{wm}
\icmlauthor{Xiaolong Ma}{neu}
\icmlauthor{Yanzhi Wang}{neu}
\icmlauthor{Bin Ren}{wm}
\end{icmlauthorlist}

\icmlaffiliation{wm}{Department of Computation, College of William \& Mary, Williamsburg, USA}
\icmlaffiliation{neu}{Department of Electrical and Computer Engineering, Northeastern University, Boston, USA}

\icmlcorrespondingauthor{Wei Niu}{wniu@email.wm.edu}
\icmlcorrespondingauthor{Bin Ren}{bren@cs.wm.edu}
\icmlcorrespondingauthor{Xiaolong Ma}{ma.xiaol@husky.neu.edu}
\icmlcorrespondingauthor{Yanzhi Wang}{yanz.wang@northeastern.edu}

% % You may provide any keywords that you
% % find helpful for describing your paper; these are used to populate
% % the "keywords" metadata in the PDF but will not be shown in the document

\vskip 0.2in
]



\printAffiliationsAndNotice{}  % leave blank if no need to mention equal contribution

\begin{abstract}
With the rapid emergence of a spectrum of high-end mobile devices, many applications that required desktop-level computation capability formerly can now run on these devices without any problem. However, without a careful optimization, executing Deep Neural Networks (a key building block of the real-time video stream processing that is the foundation of many popular applications) is still challenging, specifically, if an extremely low latency or high accuracy inference is needed. This work presents CADNN, a programming framework to efficiently execute DNN on mobile devices with the help of advanced model compression (sparsity) and a set of thorough architecture-aware optimization. The evaluation result demonstrates that CADNN outperforms all the state-of-the-art dense DNN execution frameworks like TensorFlow Lite and TVM.   
\end{abstract}

\section{Introduction}\label{sec:intro}

Nowadays, car autopilot and augmented reality (AR) techniques\footnote{\url{https://developers.google.com/glass/}, \url{https://www.microsoft.com/en-us/hololens/}} become increasingly popular, which is mainly boosted by the dramatic enhancement of real-time continuous video stream processing in the past few years~\cite{chen2015glimpse}.
Deep Neural Networks (DNN) such as Convolution Neural Networks (CNN) and Recurrent Neural Networks (RNN) serve as the state-of-the-art foundation of high-quality real-time continuous video stream processing. 
Due to its high demand for computation power and memory storage, processing video streams with DNN in real time on modern mobile devices is highly challenging. 

There have been many efforts targeting this issue, such as DeepMon~\cite{huynh2017deepmon}, DeepX~\cite{lane2016deepx}, DeepSense~\cite{yao2017deepsense}, MCDNN~\cite{han2016mcdnn}, TVM~\cite{chen2018tvm}, TensorFlow Lite~\cite{lite2017android}, etc.; however, most of them do not explore the possible optimization opportunities like computation and memory footprint reductions offered by model compression, including weight pruning \cite{han2015learning,wen2016learning,tianyunADMM1} and weight quantization \cite{zhou2017incremental,wu2016quantized,hubara2016binarized,rastegari2016xnor}. Therefore, a significant performance gap still exists between the peak performance that can be potentially offered by state-of-art mobile devices and what the existing systems actually achieved.

There are two major obstacles when leveraging model compression to improve the DNN inference execution performance in the mobile environment: first, model compression (like weight pruning/sparsity and weight quantization) usually results in accuracy degradation in inference; and second, the computation pattern of compressed models becomes more irregular, causing more severe data locality and load balancing issues and significantly increasing the difficulty in optimizations. 

Within this context, this work proposes CADNN, a programming framework to efficiently execute DNN inference on mobile devices with a more advanced model compression method that is designed to minimize the accuracy drop and maximize the model compression rate, and a set of careful {\em architecture-aware} optimizations that can effectively address the extra irregularity brought by the model compression. To our best knowledge, CADNN offers the most thorough study of optimizing compressed DNN on mobile devices and achieves the most significant performance gains compared to existing state-of-the-art dense DNN execution frameworks.     

The contribution of this work is as follows:

\begin{itemize}
    \item CADNN adopts the advanced model compression method that currently achieves the highest weight pruning rates. It can result in 348$\times$, 36$\times$, and 8$\times$ weight pruning rates with (almost) zero accuracy loss for LeNet-5, AlexNet, and ResNet-18 models, respectively. 
    \item CADNN carefully studies the performance challenges brought by model compression and offers a set of architecture-aware optimizations, such as \emph{computation pattern transformation}, \emph{redundant memory load elimination}, \emph{smart selection of memory and computation optimization parameters}, etc. 
    \item CADNN is evaluated on a modern mobile platform with its CPUs and GPUs respectively and demonstrates superior execution performance. 
\end{itemize}

Specifically, compared to the state-of-art dense DNN execution frameworks like TensorFlow Lite and TVM, CADNN can achieve up to $8.8\times$ and $6.4\times$ speedup according to our evaluation on 4 popular DNNs used in mobile applications, including Inception-V3, MobileNet, and Resnet50. %\todo[inline]{more models}
\section{Overview}

\begin{figure}[t]
    \centering
     \includegraphics[width=0.8\columnwidth]{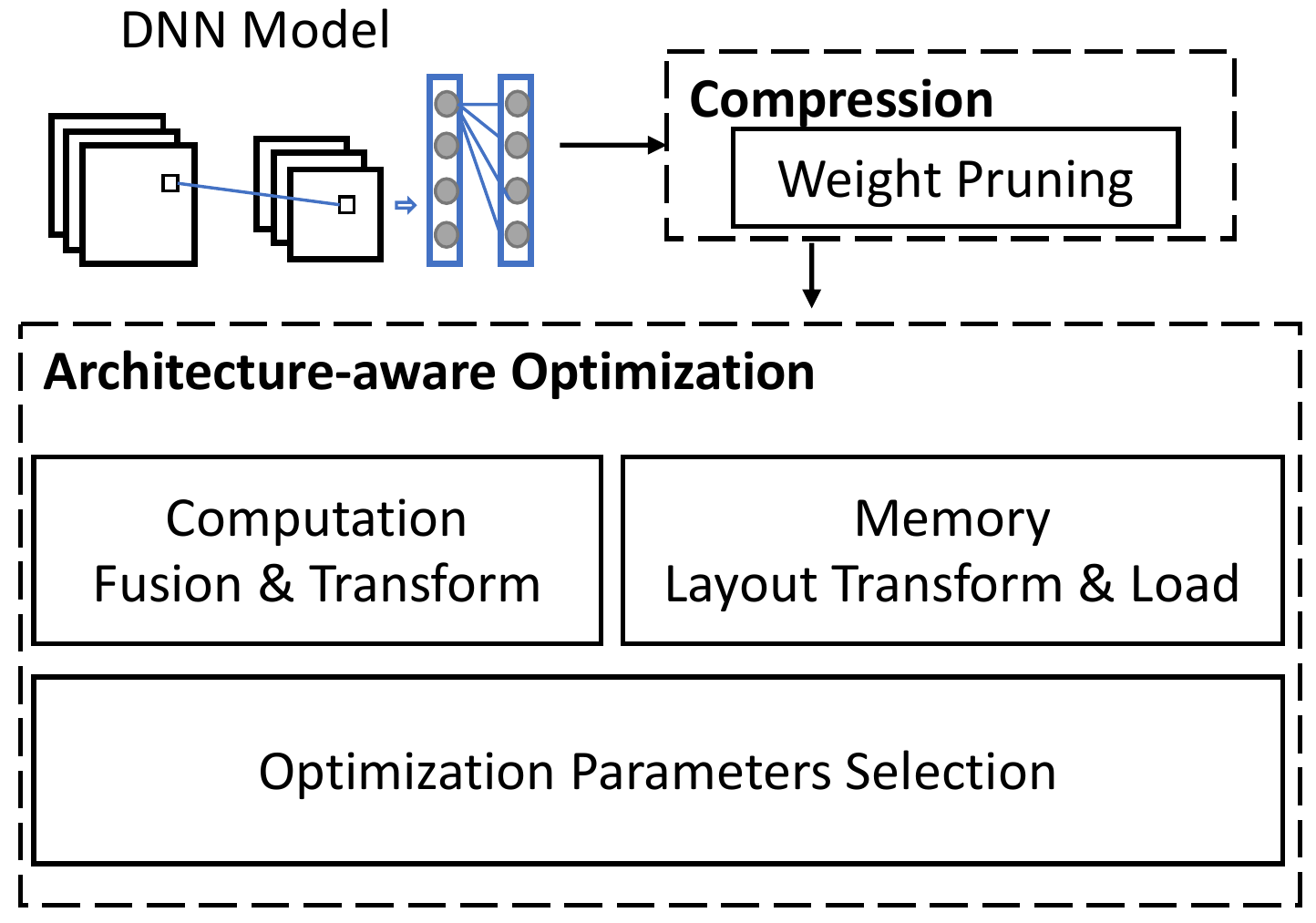}
    \caption{Overview of CADNN}\label{fig:opt-overview}
\end{figure}

Figure~\ref{fig:opt-overview} illustrates the design overview of CADNN framework that includes two stages, compression with weight pruning and a set of architecture-aware optimization. In particular, architecture-aware optimization consists of three major components, model computation fusion and transformation, memory layout transformation and load optimization, and optimization parameters selection. The basic idea of each part is explained in the following sections.
\section{Unified DNN Model Compression Framework using ADMM}

Recent work \cite{tianyunADMM1}~\cite{zhang2018adam}~\cite{ye2019progressive} has proposed a systematic DNN weight pruning technique using the advanced variable-splitting optimization method ADMM (Alternating Direction Methods of Multipliers) \cite{boyd2011distributed}. ADMM is a powerful optimization tool, by decomposing an original problem into two subproblems that can be solved separately and efficiently \cite{boyd2011distributed}. Consider optimization problem $\min_{\bf{x}}f({\bf{x}})+g({\bf{x}})$. In ADMM, it is decomposed into two subproblems on $\bf{x}$ and $\bf{z}$ (auxiliary variable), which will be solved iteratively until convergence. The first subproblem derives $\bf{x}$ given $\bf{z}$: $\min_{\bf{x}}f({\bf{x}})+q_1({\bf{x}}|{\bf{z}})$. The second subproblem derives $\bf{z}$ given $\bf{x}$: $\min_{\bf{z}}g({\bf{z}})+q_2({\bf{z}}|{\bf{x}})$. Both $q_1$ and $q_2$ are quadratic functions.

As a special property, ADMM can effectively deal with a subset of combinatorial constraints and yield optimal (or at least high quality) solutions \cite{hong2016convergence}. The associated constraints in DNN weight pruning belong to this subset of combinatorial constraints, making ADMM applicable to this specific problem. In weight pruning problem, $f(\bf{x})$ is the DNN loss function and the first subproblem is DNN training with dynamic regularization, which is totally compatible with current gradient descent techniques and solution tools for DNN training. $g(\bf{x})$ corresponds to the combinatorial constraints on the number of weights for weight pruning. As a result of the compatibility with ADMM, the second subproblem has an optimal, analytical solution via Euclidean projection. 

The ADMM-based method achieves state-of-art weight pruning results, e.g., 21$\times$ overall weight reduction in AlexNet. However, it lacks the algorithm-wise guarantee of solution feasibility, i.e., all constraints should be satisfied. We propose to extend over \cite{tianyunADMM1} in the following three aspects. First, we propose an integration of \emph{ADMM regularization} and \emph{masked mapping and retraining}. The former step resembles \cite{tianyunADMM1} while the later step guarantees solution feasibility and further improves solution quality (in terms of pruning rates). Second, we propose a unified solution framework of weight pruning and weight quantization using ADMM. For weight quantization, the combinatorial constraints are also compatible with ADMM. $g(\bf{x})$ corresponds to the combinatorial constraints on the values each weight could take, and the second subproblem again has an optimal analytical solution. Third, we also develop effective techniques, including multi-$\rho$ technique and progressive model compression, for enhancing convergence speed and quality. We release codes and models at the anonymous link:  \url{http://bit.ly/2WMQSRi}.

Our proposed framework achieves the best-in-class weight pruning rate, as well as for weight pruning combined with quantization. For non-structured weight pruning alone, we achieve 348$\times$ overall weight reduction (only 0.28\% remaining weights) in LeNet-5, 36$\times$ in AlexNet (ImageNet dataset), 34$\times$ for VGGNet, and 9.2$\times$ on ResNet-50, with (almost) no accuracy loss, 2$\times$ to 28$\times$ improvement over competing methods. We achieve a reduction of up to 3,438$\times$ in weight storage (using LeNet-5 model, not accounting for indices), with almost no accuracy loss when weight pruning and quantization are combined, outperforming the state-of-art by two orders of magnitude.

\section{Architecture-aware Optimization}

CADNN supports three major optimizations targeting modern mobile architectures as follows: 

\noindent{\bf Model computation fusion and transformation} 
After weight pruning, each layer's computation is significantly reduced; however, their memory access becomes much more irregular. This further magnifies the {\em memory wall} issue of mobile devices, rendering the execution increasingly memory-bound. CADNN therefore explores every opportunity in DNN to fuse multiple layers into larger computation kernels (e.g. Convolution layer/Depthwise Convolution layer $+$ BatchNorm layer $+$ Activation layer in MobileNetV1), yielding benefits in two aspects: first, reducing the intermediate irregular memory read and write therefore improving memory performance; and second, packing more computation workloads together therefore increasing the SIMD (Single-Instruction Multiple Data) utilization. In particular, for convolution layers with $1\times1$ filters, CADNN is able to further transform the convolution operation into matrix multiplication operation to further improve its memory and SIMD performance.    

\noindent{\bf Memory layout transformation and load optimization}
To further improve the data locality, CADNN also transforms the memory layout of DNN's filters to fit CPU and GPU, respectively. More specific techniques include tiling, alignment, and padding. Specifically, based on a key observation that many elements in filters of convolution layers are repeatedly loaded to registers, CADNN implements a compiler code transformation to eliminate such redundant memory loads.   

\noindent{\bf Optimization parameters selection}
Above optimizations work for varied DNNs on both CPU and GPU, however, with very different parameters, e.g., for a specific DNN, the best tile sizes for CPU and GPU are different from each other, and the best tile sizes for varied layers are also different. A large number of optimization parameters (such as tiling sizes on multiple dimensions, unrolling sizes, possible computation reorders, etc.) imposes a big challenge for us to select the best configuration. CADNN adopts an optimized tuning approach by pruning the redundant or sub-optimal configurations with the knowledge from both DNNs and architectures, and then uses a compiler source-to-source code transformation to generate optimized computation kernels.    
\section{Evaluation}

We compare CADNN's performance with TensorFlow Lite (a popular framework on the mobile device), and TVM (the state-of-the-art framework).
%\noindent{\bf Objective:}
\begin{figure}[h]
    \centering
     \includegraphics[width=1.0\columnwidth]{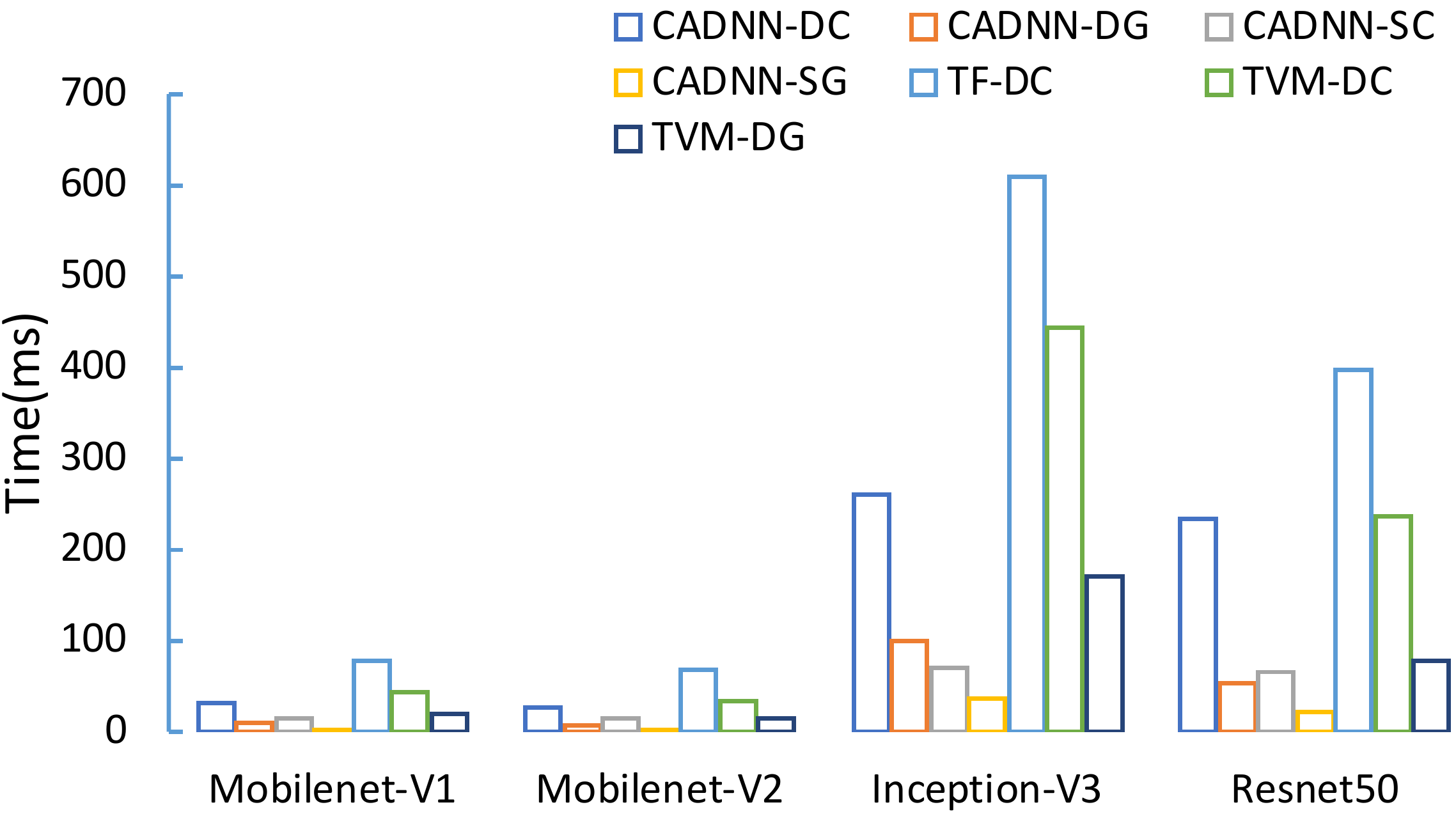}
     \vspace{-1.5em}
    \caption{Inference performance comparison on CPU/GPU: CADNN-DC (CADNN dense on CPU), CADNN-DG (CADNN dense on GPU), CADNN-SC (CADNN compressed on CPU), CADNN-SG (CADNN compressed on GPU), TFLITE-DC (TensorFlow Lite dense on CPU), TVM-DC (TVM dense on CPU), and TVM-DG (TVM dense on GPU) }\label{fig:performance}
\end{figure}

\noindent{\bf Platform} CADNN is evaluated on a Xiaomi 6 cell phone whose detailed configuration is summarized in Table~\ref{tab:device-info}.  
\begin{table}[ht]
\caption{Device information of Xiaomi 6 with Android 8.0.0}
{\small
\begin{center}
\begin{tabular}{|c||c|}
\hline
\textbf{SOC} & Snapdragon 835, up to 2.45GHz\\
\hline
\textbf{GPU} & Adreno 540, 710MHz\\
\hline
\textbf{Memory} & 6GB shared by CPU and other devices\\
\hline
\end{tabular}
\label{tab:device-info}
\end{center}
}
\end{table}

\begin{table}[t]
\caption{DNN Configurations}
\small{
\begin{center}
\begin{tabular}{|l|l|l|l|l|}
\hline
\textbf{Model} & \textbf{Size(M)} & \textbf{Top1 (\%)} & \textbf{Top5 (\%)} & \textbf{Layer} \\ \hline
MobileNet-V1   & 17.1             & 70.9                        & 89.9                        &31                 \\ \hline
MobileNet-V2   & 14.1             & 71.9                        & 91.0                        &66                 \\ \hline
Inception-V3   & 95.4             & 78.0                        & 93.9                        &126                 \\ \hline
Resnet50       & 102.4            & 75.2                        & 92.2                        &94                 \\ \hline
%VGG-16         & 553.4            & 71.5                        & 89.8                        &                 \\ \hline
\end{tabular}
\label{tab:dnn-config}
\end{center}
}
\end{table}

\noindent{\bf DNNs and data-set} Table~\ref{tab:dnn-config} characterizes the four DNNs used in our evaluation, showing the model name, size, top-1 accuracy, top-5 accuracy, and the number of layers. All tests are performed on the ImageNet data-set.

\noindent{\bf Performance} Figure~\ref{fig:performance} shows the performance comparison between CADNN and TensorFlow Lite/TVM. CADNN supports both dense and compressed models while TensorFlow Lite and TVM support only dense models. We show CADNN and TVM's performance on both CPU and GPU, and TensorFlow Lite on CPU only. On both CPU and GPU, CADNN outperforms the other two, achieving up to $6.4\times$ and $6\times$ speedup over TVM (that is better than TensorFlow Lite) on CPU and GPU, respectively. This result also demonstrates CADNN is super fast, and with relatively large DNNs, Inception-V3 and Resnet50, it can finish inference of a single image within 35 ms and 21 ms, respectively.

%\subsection{Performance on GPU}

% \begin{table}[htbp]
% \caption{Device information of Xiaomi 6 with Android 8.0.0}
% \begin{center}
% \begin{tabular}{|c|c|c|}
% \hline
% \cline{2-2} 
% \textbf{\textit{SOC}}& \textbf{\textit{GPU}}& \textbf{\textit{Memory}} \\
% \hline
% Snapdragon 835, \\ up to 2.45GHz & Adreno 540, 710MHz & 6GB\\
% \hline
% \end{tabular}
% \label{tab3}
% \end{center}
% \end{table}

\section{Conclusion and Work in Progress}\label{sec:conclusion}

This work presents CADNN, a programming framework to efficiently execute DNN inference on mobile devices with the help of a more advanced model compression and a set of architecture-aware optimizations. Our evaluation shows that CADNN is super fast, achieving up to $8.8\times$ and $6.4\times$ speedup over TensorFlow Lite and TVM, two popular and highly optimized dense DNN execution frameworks. We plan to further optimize CADNN in two aspects: a smarter optimization parameters selection scheme based on other Machine Learning techniques, and a DNN profiler on mobile devices to better detect the performance bottleneck of DNN execution.


% \bibliography{ref}
% \bibliographystyle{icml2019}

% \appendix
% \section{Do \emph{not} have an appendix here}

% \textbf{\emph{Do not put content after the references.}}
% %
% Put anything that you might normally include after the references in a separate
% supplementary file.

% We recommend that you build supplementary material in a separate document.
% If you must create one PDF and cut it up, please be careful to use a tool that
% doesn't alter the margins, and that doesn't aggressively rewrite the PDF file.
% pdftk usually works fine. 

% \textbf{Please do not use Apple's preview to cut off supplementary material.} In
% previous years it has altered margins, and created headaches at the camera-ready
% stage. 
%%%%%%%%%%%%%%%%%%%%%%%%%%%%%%%%%%%%%%%%%%%%%%%%%%%%%%%%%%%%%%%%%%%%%%%%%%%%%%%
%%%%%%%%%%%%%%%%%%%%%%%%%%%%%%%%%%%%%%%%%%%%%%%%%%%%%%%%%%%%%%%%%%%%%%%%%%%%%%%

\end{document}